\pdfoutput=1
\documentclass[11pt]{article}

\usepackage[preprint]{acl}
\usepackage{float}
\usepackage{times}
\usepackage{latexsym}
\usepackage{amsmath}
\usepackage[T1]{fontenc}
\usepackage[utf8]{inputenc}

\usepackage{microtype}

\usepackage{inconsolata}

\usepackage{graphicx}

\title{One Jump Is All You Need: Short-Cutting Transformers for Early Exit Prediction with One Jump to Fit All Exit Levels}

\author{Amrit Diggavi Seshadri \\
    Sudarshantech Software \\
  \texttt{amrit@sudarshantechsoftware.com}}

\begin{document}
\maketitle
\begin{abstract}
To reduce the time and computational costs of inference of large language models, there has been interest in parameter-efficient low-rank early-exit casting of transformer hidden-representations to final-representations. Such low-rank short-cutting has been shown to outperform identity shortcuts at early model stages while offering parameter-efficiency in shortcut jumps. However, current low-rank methods maintain a separate early-exit shortcut jump to final-representations for each transformer intermediate block-level during inference. In this work, we propose selection of a single One-Jump-Fits-All (OJFA) low-rank shortcut that offers over a 30x reduction in shortcut parameter costs during inference. We show that despite this extreme reduction, our OJFA choice largely matches the performance of maintaining multiple shortcut jumps during inference and offers stable precision from all transformer block-levels for GPT2-XL, Phi3-Mini and Llama2-7B transformer models.
\end{abstract}
\section{Introduction}
Large Language Models \cite{vaswani2017attention} stack multiple transformer-blocks of multi-headed self-attention and feed forward layers sequentially, with modern models stacking upwards of 30 such blocks. For example GPT2-XL \cite{radford2019language} stacks 48 such blocks, Phi3-Mini \cite{abdin2024phi} and Llama2-7B \cite{touvron2023llama} each stack 32 blocks, while larger models like GPT-3 \cite{brown2020language} and  Llama3-405B \cite{dubey2024llama} stack as many as 96 and 126 transformer-blocks. 
\begin{figure}[!h]
    \centering
    \includegraphics[width=\linewidth]{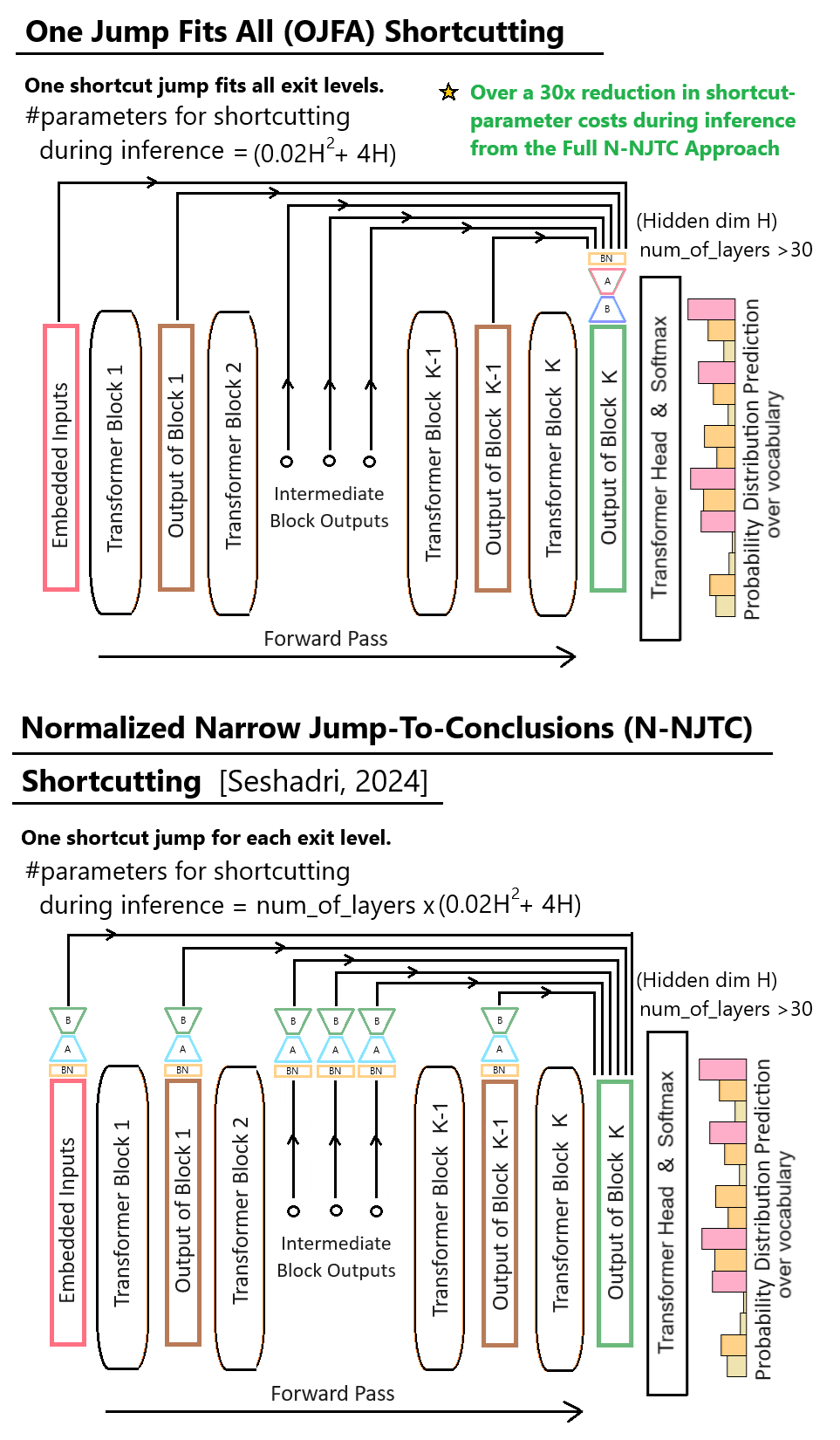}
    \caption{Illustration of our One-Jump-Fits-All (OJFA) approach in comparison to the previous method.}
    \label{fig:depict}
\end{figure}
Inputs are forward passed through these transformer-blocks sequentially to obtain increasingly sophisticated sentence-contextualized token-representations, and while intermediate token-representations are created at each transformer block-level, normally only the final representations are used by the transformer model-head for output-prediction. 
In contrast to the traditional approach, attempts have been made to shortcut-cast token-representations from intermediate transformer block-levels to final-representations during model-inference \cite{schwartz2020right, geva2022transformer, din2023jump, seshadri-2024-normalized}, with the condition to exit the forward-pass once intermediate-prediction has reached some preset confidence-level $\lambda$ for faster and cheaper inference.
In particular, N-NJTC short-cut jumps \cite{seshadri-2024-normalized} that cast token-representations to final representations using low-rank linear matrices has recently been shown to consistently outperform identity shortcuts at early transformer-block levels while being significantly cheaper in parameter costs than previous full-rank methods \cite{din2023jump}. However, in order to allow an early-exit option from each block level, even the parameter-efficient N-NJTC approach requires that we maintain a separate low-rank shortcut jump for each early-exit block-level during inference (Fig. \ref{fig:depict}). For modern models with over 30 such block-levels, this means maintaining over 30 separate shortcut-jumps during inference. With the size and depth of transformer models increasing, clearly there is a need to curtail this growing shortcut parameter cost. In this work, we demonstrate that it is in fact possible to replace multiple low-rank shortcut jumps by a single well-chosen low-rank shortcut jump while still largely retaining performance at all early-exit levels.
\begin{itemize}
    \item We propose selection of a  single One-Jump-Fits-All (OJFA) low-rank shortcut that offers over a 30x reduction in total shortcut parameter costs from the previous multi-jump N-NJTC method during ealy-exit inference.
    \item We show that despite this extreme parameter reduction, our OJFA choice largely matches the performance of maintaining multiple shortcut jumps in N-NJTC during inference and offers stable precision for early exit from all transformer block-levels for GPT-2XL, Phi3-Mini and Llama2-7B transformer models.
\end{itemize}
\section{Related Work}
\cite{schwartz2020right} was the first to propose Identity shortcuts - that feed transformer token-representations from intermediate block-levels to the model head directly without modification. \cite{din2023jump} considered the output of different transformer blocks to operate in different representational spaces and demonstrated that linear matrix shortcut jumps (JTC shortcuts) greatly outperform Identity shortcutting. However, JTC is computationally expensive. \cite{seshadri-2024-normalized} proposed N-NJTC low-rank normalized linear shortcuts that outperformed Identity shortcuts at early model stages while using less than $3\%$ the number of parameters of a JTC shortcut. However, as mentioned earlier, N-NJTC is still a multi-jump method that maintains a separate shortcut for each transformer block-level at inference. In this work, we use the low-rank shortcut mechanism proposed by N-NJTC, but make a careful selection of a single-shortcut to fit all exit levels. To our knowledge, we are the first to apply a single shortcut jump to multiple exit-levels in transformer shortcutting.
\section{Method}
\label{method}
For a given set of $N$ input sentences $S$ we forward pass each sentence $s_i \in S$ through a given transformer to obtain intermediate-block-representations $\{h^k_i\}_{i=1}^N$ at each block-level $k$ and final-representations $\{h^K_i\}_{i=1}^N$ for randomly selected token-positions in each $s_i$.
As mentioned earlier, we use the low-rank N-NJTC shortcut jump mechanism \cite{seshadri-2024-normalized}, which consists of 2 low-rank linear matrices $A_k$ and $B_k$ preceded by a batch-normazlization layer\footnote{Where we have $A_k: H \times \lfloor\frac{H}{100}\rfloor$ and $B_k: \lfloor\frac{H}{100}\rfloor \times H$ for transformer of hidden dimension $H$; and $K$ exit-block-levels}. A shortcut is trained to minimize the MSE-loss $L_{k}$ between approximated final-representations $\hat{h}^{kK}_i$ and true transformer final-representations $h^K_i$ at each level $k$.
\begin{figure*}[!h]
    \centering
    \includegraphics[width=\linewidth]{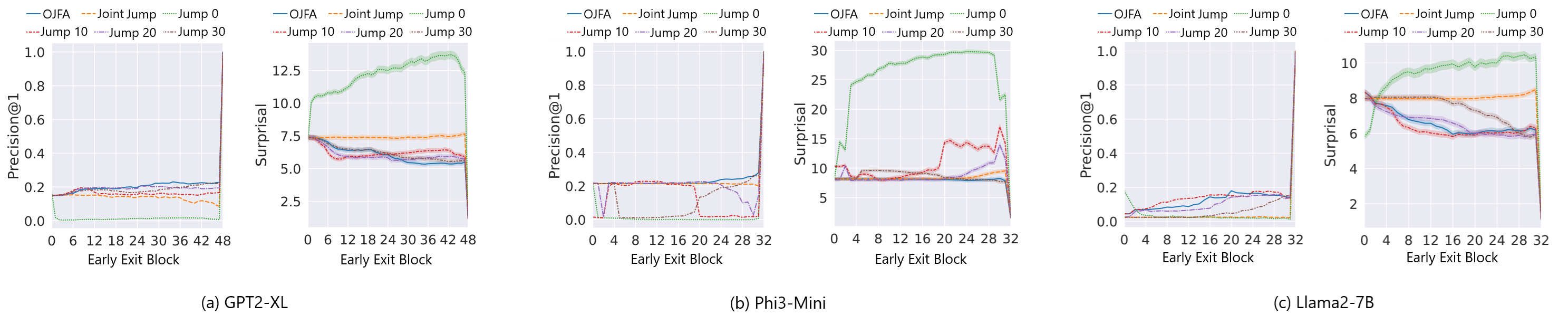}
    \caption{Precision ($\uparrow$) and Surprisal ($\downarrow$) using our OJFA-choice shortcut-jump for early-exit prediction in comparison to arbitrary-choice jumps and jointly-trained shortcut-jumps for (a) GPT2-XL, (b) Phi3-Mini and (c) Llama2-7B.}
    \label{fig:arbitrary}
\end{figure*}
\begin{equation}
\label{eq-1}
    \hat{h}^{kK}_i = \text{BatchNorm}_k(h^k_i)A_kB_k
\end{equation}
\begin{equation}
\label{eq-2}
    L_{k} = \frac{1}{N}\sum_{i=1}^{N} ||\hat{h}^{kK}_i - h^K_i||^2 
\end{equation}
\subsection{One Jump Fits All (OJFA) Choice}
As mentioned earlier, in this work, we make an automated choice of a single OJFA shortcut-jump from one of the early-exit levels $\{0,..,K-1\}$ that can be reused by all exit levels at inference. As shown below, we denote $\hat{h}^{kKm}_i$ to be the approximated final-representation from intermediate-block-level $k$ computed using shortcut-parameters $\{A_m, B_m, \text{BatchNorm}_m\}$ from jump $m$ that have been trained to minimize $L_m$ (Eq.\ref{eq-1} \& \ref{eq-2}.)
\begin{equation}
\label{eq-3}
    \hat{h}^{kKm}_i = \text{BatchNorm}_m(h^k_i)A_mB_m
\end{equation}
To make our OJFA choice of jump-parameters in a task-agnostic way (independent of the transformer model-head), we consider minimizing a Signed Sensitive Cosine Similarity measure $D_m$ as shown below.
\begin{equation}
    C_i^{km} =  \frac{\hat{h}^{kKm}_i \cdot h^K_i}{||\hat{h}^{kKm}_i||\cdot||h^K_i||}
\end{equation}
\begin{equation}
    \label{sscs}
    D_m = \frac{1}{NK}\sum_{i=1}^{N}\sum_{k=1}^{K}\text{sgn}(C_i^{km}) \left(C_i^{km}\right)^{2}
\end{equation}
Concretely, cosine similarity $C_i^{km}$ between final-approximations $\hat{h}^{kKm}_i$ (arrived at from an exit-level $k$ using jump-parameters from shortcut $m$), and true-final-representations $h^K_i$ provides a direct measure of correctness. We consider the signed square of this value to make this measure of correctness more sensitive to very correct and very incorrect approximations, while placing less emphasis on orthogonal inaccurate predictions, and sum over all exit-levels, and all data-points in the training data to arrive at a single score  $D_m$ that measures the efficacy of exit-jump $m$ when reused at all exit levels. Our OJFA choice is then the choice of the shortcut-jump that maximizes the $D_m$ score for $m \in \{0,..,K-1\}$. All other shortcut-jump parameters besides the OJFA choice can then be discarded at inference as we use only our OJFA choice parameters at all early-exit levels according to Eq.\ref{eq-3}. As discussed earlier, modern transformer models consist of $K>30$ block-levels, and reusing a single shortcut-jump for all levels at inference allows us to make over a 30x reduction in shortcut-parameter costs over the previous multi-jump state-of-the-art.
\section{Experiments}
\iffalse
\begin{figure}[!h]
    \centering
\includegraphics[width=\linewidth]{correlation_scores.png}
    \caption{Caption}
    \label{fig:enter-label}
\end{figure}
\fi
We test our OJFA-choice on GPT2-Xl \cite{radford2019language} that consists of 48 transformer-blocks and hidden dimension 1600 - for which OJFA allows a 48x reduction in shortcut-parameter costs during inference - reducing total shortcut-parameter count from 2.76M down to 57.6K parameters at inference; on the larger Phi3-Mini \cite{abdin2024phi} that consists of 32 transformer blocks and hidden dimension size 3072 - for which OJFA allows a 32x reduction in shortcut-parameter costs during inference - reducing total shortcut-parameter count from 6.29M down to 196.6K parameters at inference; and on the even larger Llama2-7B \cite{touvron2023llama} that consits of 32 transformer blcoks and hidden dimension size 4096 - for which OJFA allows a 32x reduction in shortcut-parameter costs during inference - reducing total shortcut-parameter count from 11.01M down to 344.06K parameters at inference.
\\\\
\textbf{Data:} Following the approach taken by \cite{din2023jump, seshadri-2024-normalized}, we random sample 9000 train sentences and 3000 test sentences from Wikipedia - each of which are highly diverse, written by different authors on different topics. As explained in Section \ref{method}, we forward pass each sentence through a given transformer model and use random token position representations to train and evaluate shortcutting approaches.
\begin{figure}[!h]
    \centering
\includegraphics[width=\linewidth]{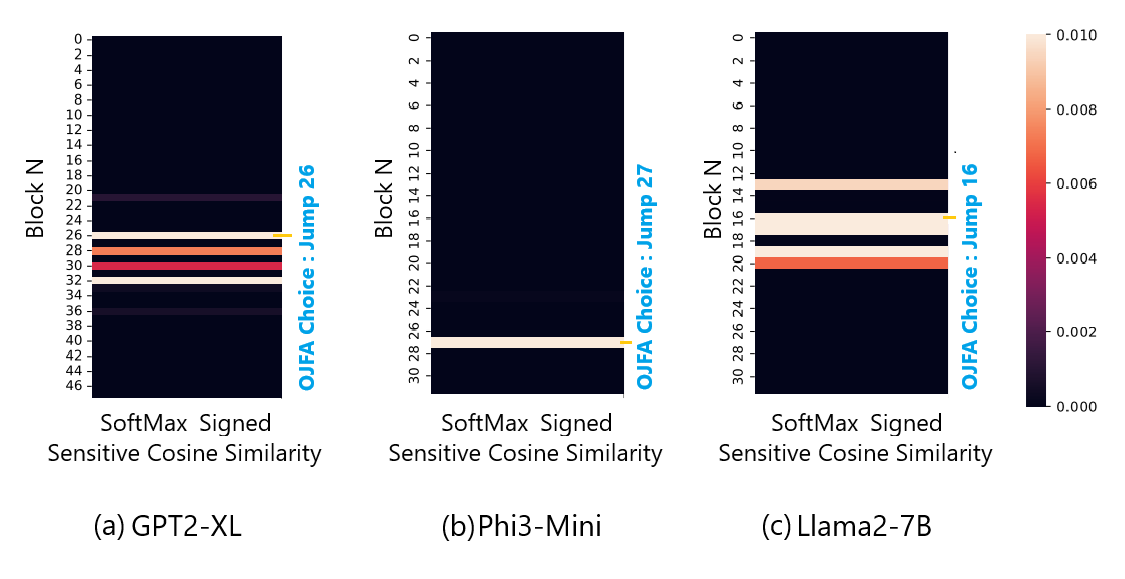}
    \caption{SoftMax Signed Sensitive Cosine Similarity scores for each Shortcut-Jump from Block N to final-representations, showing selection of our OJFA choice.}
    \label{fig:softmax}
\end{figure}
\section{Comparison with Arbitrary Choice and Joint Shortcut Training}
We first make our OJFA-choices of shortcut-jump for each model to maximize Signed Sensitive Cosine Similarity (Eq.\ref{sscs}). Figure \ref{fig:softmax} shows a SoftMax distribution of these scores.\footnote{At low temperature $5e-4$ to create a sharp distribution.}
\begin{figure}[!h]
    \centering
    \includegraphics[width=\linewidth]{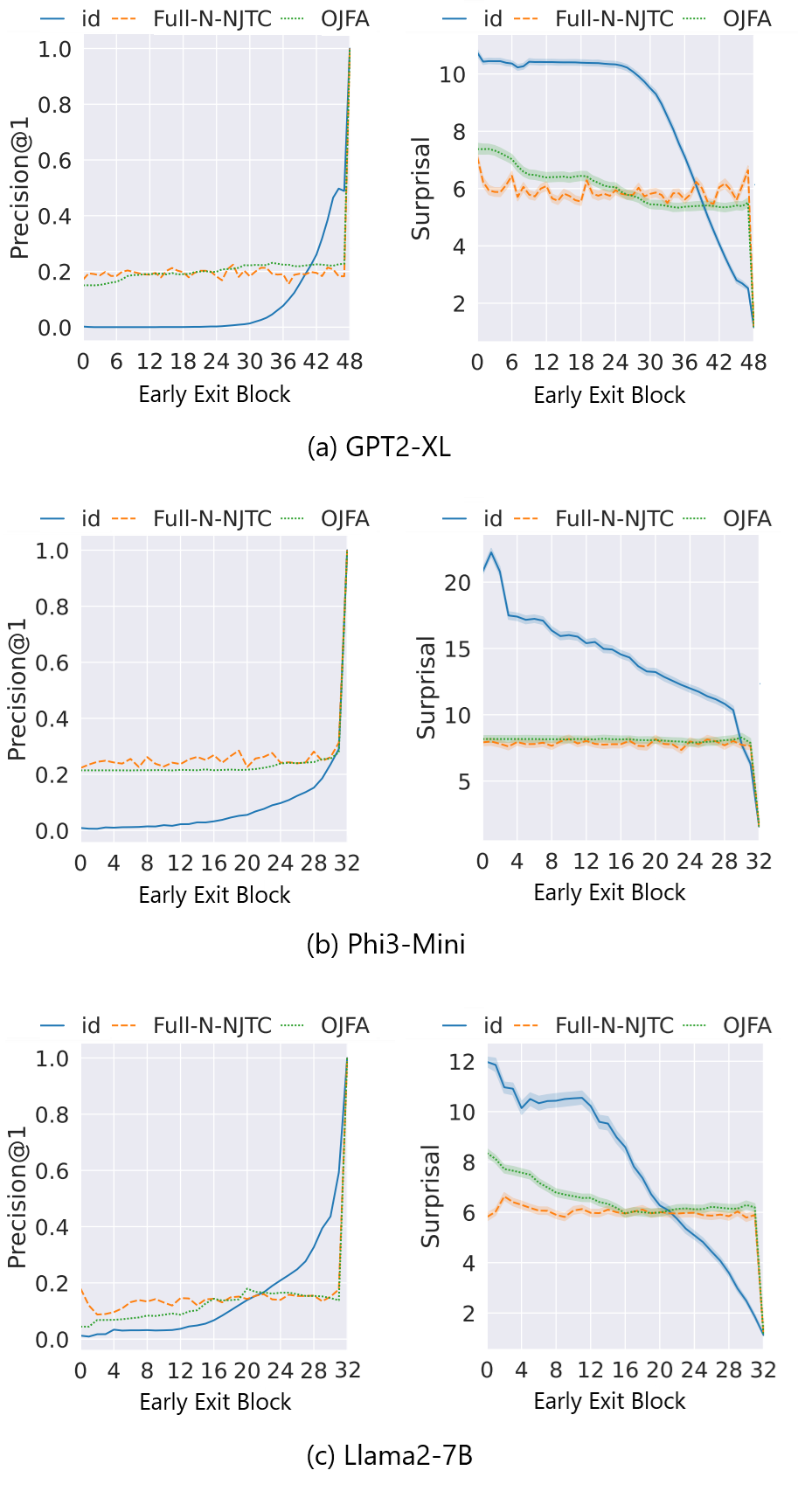}
    \caption{Precision ($\uparrow$) and Surprisal ($\downarrow$) achieved by our single OJFA-choice jump for early-exit prediction at all exit-levels in comparison to Identity Shortcuts and the Full N-NJTC with a separate jump for each exit-level for (a) GPT2-XL, (b) Phi3-Mini and (c) Llama2-7B.}
    \label{fig:final}
\end{figure}
As shown in the figure, our OJFA-choices are jump 26, jump 27, and jump 16 for GPT2-XL, Phi3-Mini and Llama2-7B models respectively, i.e the shortcut-parameters from block exit-levels 26, 27 and 16 to final representations in each model. With these choices made, we evaluate the efficacy of our OJFA-choices on the task of next token prediction. Specifically, we compare using our OJFA-choice at all exit-levels to using arbitrary-choice jumps at all exit-levels and a custom trained joint-jump shortcut that is trained to minimize MSE with true-final-representations (Eq.\ref{eq-2}) for intermediate-representation inputs sourced from all exit-levels\footnote{We evaluate scores at each level using fixed-early-exits from the transformer's forward-pass, which is a stricter evaluation criteria than heuristic-informed early-exit.}.
Following the approach taken by \cite{din2023jump, seshadri-2024-normalized}, we compute Precision by assigning a score of 1 if the most-likely next-token from the shortcut-predicted distribution matches the most-likely next-token from the true-final-predicted distribution and 0 otherwise. Surprisal is computed as the negative log-likelihood of the true-final-predicted distribution's most likely output token according to the shortcut predicted distribution. As shown in Figure \ref{fig:arbitrary}, on average, for all exit-levels, our OJFA choice achieves better precision and surprisal scores than using arbitrary choices of shortcut-jumps - arbitrary choices of shortcuts do not always generalize well to all levels of early-exit prediction. Importantly, we also find that it is better to use our single OJFA-choice of a shortcut jump that is trained for one-exit level for early-exit prediction at all exit-levels rather than train a custom joint-jump that has been specifically trained to accept intermediate-representation inputs from any exit-level - a possible explanation for this behavior is that such joint-training presents a very noisy problem setting. The structure of noise or extra encoded information in early-representations is likely easier for shortcut parameters to identify and ignore when training at a single exit level than it is to identity when the training inputs are sourced from multiple-exit-levels and are more diverse.

\section{Comparison with Identity Shortcuts and Full Multi-Jump N-NJTC}
Finally, we make a comparison between our single OJFA-choice, Identity Shortcuts (id) \cite{schwartz2020right}, and the Full Multi-Jump N-NJTC approach \cite{seshadri-2024-normalized} at all exit-levels. As shown in Figure \ref{fig:final}, we find that despite the extreme parameter-reduction made by using our OJFA-choice, we still outperform id shortcuts at early model stages and are largely able to match the performance of the full N-NJTC approach at all exit-levels for GPT2-XL, Phi3-Mini and Llama2-7B. 
\section{Conclusion}
In this work, we proposed selection of a single One-Jump-Fits-All (OJFA) low-rank shortcut that offers over a 30x reduction in shortcut-parameter costs from the previous multi-jump approach. We showed that despite this extreme reduction, OJFA still largely matches the performance of maintaining multiple low-rank shortcuts during early-exit inference from all exit-levels for GPT2-XL, Phi3-Mini and Llama2-7B transformer models. To our knowledge, we are the first to demonstrate the surprising viability of applying a single shortcut jump to multiple exit-levels in transformer shortcutting.
\section{Limitations}
Notably, low-rank shortcutting of transformers is still a relatively nascent field, and neither our method nor the previous methods are able to achieve over 60\% precision for hard early exits at any block-level for any transformer model in this parameter-efficient setting, nor is low rank-shortcutting able to outperform Identity Shortcuts in late-block shortcutting - leaving much scope for future improvement. We also find that our OJFA-choice shortcut achieves slightly worse precision and surprisal scores than the Full Multi-Jump N-NJTC approach (Figure \ref{fig:final}) at some exit-levels. We highlight however that our OJFA-choice largely matches the performance of using the Full N-NJTC method and the relatively small drop in performance observed for our method is acceptable in exchange for the over 30x reduction in parameter-cost that our approach offers during inference. In practical terms, low-rank shortcutting of transformers for early-exit prediction is already a method for extreme parameter-efficiency, skipping billions of parameter computations between intermediate and final transformer representations while compressing information down to very low-rank dimensions ($16$, $30$ and $40$ for GPT-2XL, Phi3-Mini and Llama2-7B respectively \cite{seshadri-2024-normalized}). In this context, we highlight that it is indeed very surprising that any reuse of shortcut parameters across more than one exit-level is at all possible. Finally, early-exiting transformers in general can cause unexpected model behavior, and we recommend that any shortcut inferred predictions be tested for safety before use in industrial application domains.
\bibliography{custom}
\end{document}